\documentclass[conference]{IEEEtran}
\IEEEoverridecommandlockouts
\usepackage{cite}
\usepackage{amsmath,amssymb,amsfonts}
\usepackage{algorithm}
\usepackage{algorithmic}
\usepackage{graphicx}
\usepackage{textcomp}
\usepackage{xcolor}
\usepackage{booktabs}
\usepackage{multirow}
\usepackage{caption}
\usepackage{cuted}
\usepackage{makecell}
\usepackage[normalem]{ulem}
\def\BibTeX{{\rm B\kern-.05em{\sc i\kern-.025em b}\kern-.08em
    T\kern-.1667em\lower.7ex\hbox{E}\kern-.125emX}}
\begin{document}

\title{PhyRPR: Training-Free Physics-Constrained Video Generation}

\author{
\IEEEauthorblockN{
Yibo Zhao$^{1}$\quad
Hengjia Li$^{1}$\quad
Xiaofei He$^{1}$\quad
Boxi Wu$^{1}$
}
\IEEEauthorblockA{
$^{1}$State Key Lab of CAD\&CG, Zhejiang University
}
}

\maketitle

\begin{strip}
\centering
\includegraphics[width=\textwidth]{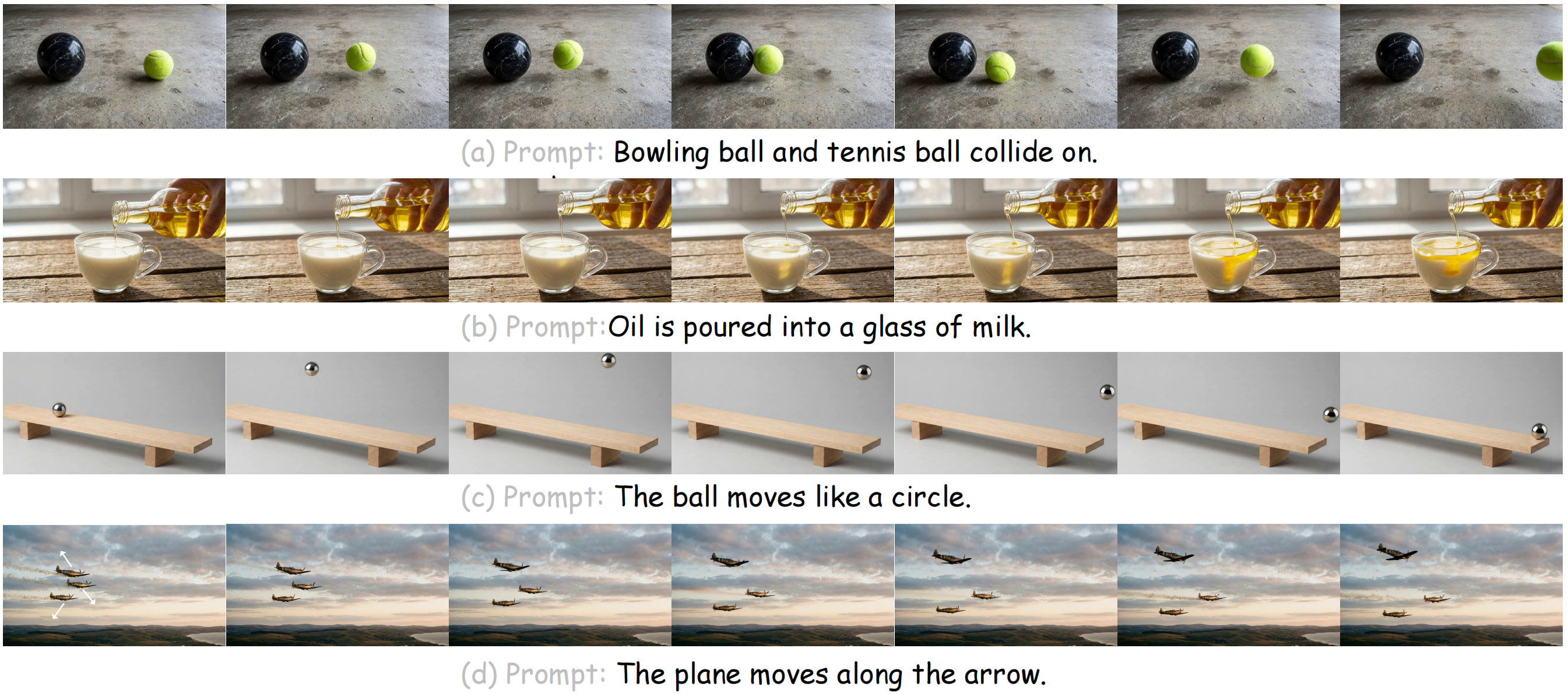}
\captionof{figure}{Samples produced by our method. (a--b) require physical priors, while (c--d) emphasize strict motion constraints. Our three-stage pipeline \textit{PhyRPR} (\textit{PhyReason}$\rightarrow$\textit{PhyPlan}$\rightarrow$\textit{PhyRefine}) decouples physical reasoning from rendering to better satisfy physical constraints while maintaining high-fidelity video quality.}
\label{fig:toutu}
\vspace{-0.6em}
\end{strip}

\begin{abstract}
    Recent diffusion-based video generation models can synthesize visually plausible videos, yet they often struggle to satisfy physical constraints.
    A key reason is that most existing approaches remain single-stage: they entangle high-level physical understanding with low-level visual synthesis, making it hard to generate content that require explicit physical reasoning.
    To address this limitation, we propose a training-free three-stage pipeline, \textit{PhyRPR}:
    \textit{Phy\uline{R}eason}--\textit{Phy\uline{P}lan}--\textit{Phy\uline{R}efine},
    which decouples physical understanding from visual synthesis.
    Specifically, \textit{PhyReason} uses a large multimodal model for physical state reasoning and an image generator for keyframe synthesis;
    \textit{PhyPlan} deterministically synthesizes a controllable coarse motion scaffold;
    and \textit{PhyRefine} injects this scaffold into diffusion sampling via a latent fusion strategy to refine appearance while preserving the planned dynamics.
    This staged design enables explicit physical control during generation.
    Extensive experiments under physics constraints show that our method consistently 
    improves physical plausibility and motion controllability.
\end{abstract} 

\begin{IEEEkeywords}
Diffusion Model, Reasoning video generation
\end{IEEEkeywords}

\section{Introduction}
\label{sec:intro}

\begin{figure}[t]
    \centering
    \includegraphics[width=\columnwidth]{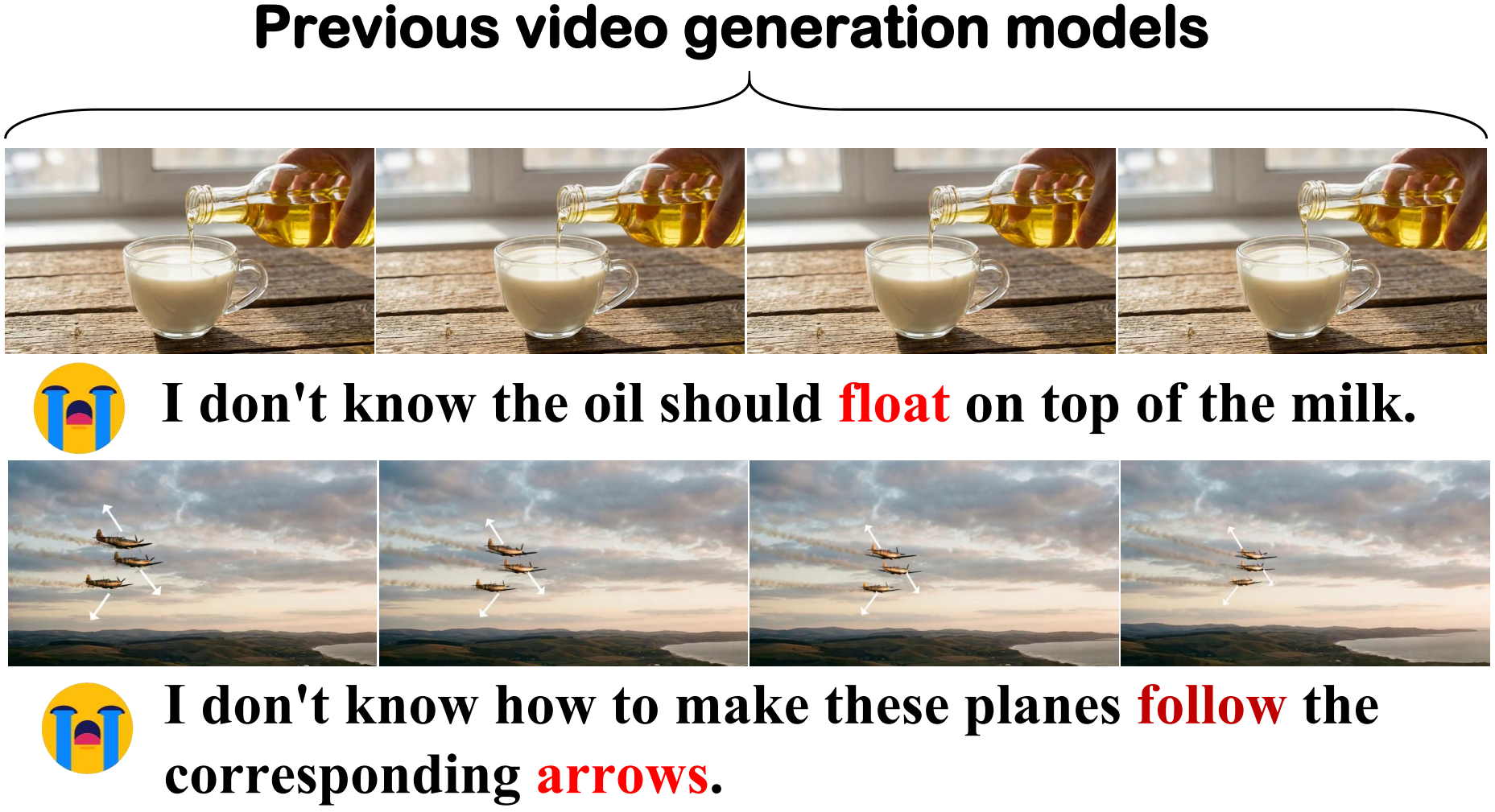}
    \caption{Prior video generation models fail to accurately follow the provided physical constraints.    }
    \label{fig:xiaotu}
    \end{figure}

In recent years, diffusion-based video generation models have made remarkable progress, synthesizing high-fidelity and visually compelling, film-like content. 
However, despite their impressive visual realism, video diffusion models remain largely correlation-driven:
they primarily exploit patterns in large-scale training data, rather than explicitly enforcing physical constraints.
As a result, they often fail in scenarios with clear physical constraints.
For example, in Fig.~\ref{fig:xiaotu}, prior models fail to (1) keep oil floating on milk and (2) follow explicit directional cues.
Since single-stage denoising does not explicitly model these constraints, they are often violated in practice, leading to inconsistent interactions and motion drift that undermine controllability and usability.

In image generation, a similar challenge arises: how to infer the deeper physical implications of a prompt through reasoning.
Prior methods address this via prompt enhancement or joint training that couples vision--language models with diffusion models.
While effective for images in some cases, directly extending these strategies to video for physically consistent generation is often insufficient.
Prompt enhancement typically targets surface-level appearance and is too imprecise to encode temporal dynamics.
Training-based video approaches are also extremely expensive, and without dedicated physics-aware annotations, models struggle to acquire generalizable physical commonsense through implicit learning.
Consequently, many video generators still rely on a single-stage denoising process where physical reasoning is only learned implicitly and entangled with visual synthesis.
This entanglement makes the generation difficult to control when explicit kinematic constraints or grounded interactions are required.

To address this problem, we propose a training-free framework that decouples physical understanding from visual synthesis via a three-stage pipeline \textit{PhyRPR}:
\textit{PhyReason}--\textit{PhyPlan}--\textit{PhyRefine}.
The key idea is to separate \emph{what should happen} from \emph{how it is rendered} by exposing intermediate, physically meaningful representations.
In \textit{PhyReason}, we leverage a large multimodal model to infer the underlying physical implications of the prompt and extract a sequence of physically grounded key states.
These states are visualized as semantically consistent keyframes, together with object-centric masks that provide explicit handles for subsequent control.
In \textit{PhyPlan}, we translate the discrete key states into continuous motion trajectories and synthesize a coarse motion scaffold that explicitly encodes object dynamics and interactions.
In \textit{PhyRefine}, we integrate this scaffold into diffusion sampling through motion-aware noise-consistent latent fusion, so that the video model can refine textures and details while staying aligned with the planned dynamics.
Overall, this staged design provides an interpretable and controllable generation process that better satisfies kinematic constraints and physics-grounded interactions.

By combining LMM-based physical priors, deterministic planning tools, and the strong rendering capability of video diffusion models, our method forms a physically interpretable, training-free video generation framework. It enables physically plausible generation as well as precise motion control. Extensive experiments demonstrate that our approach achieves clear improvements over existing methods in physical consistency, trajectory controllability, and overall visual quality.

\section{Related Work}
\subsection{Physics video generation}
Recent video generation models, like WanX\cite{wan2025wan}, Hunyuanvideo\cite{kong2024hunyuanvideo}, Cosmos\cite{ali2025world} have demonstrated impressive generative capabilities, emerging as robust rendering backends for high-fidelity synthesis. 
However, they fundamentally function as statistical mimics lacking intrinsic physical understanding. This limitation is starkly highlighted by physics-centric benchmarks such as PhyGenBench\cite{meng2024towards} and VideoScience-Bench\cite{hu2025benchmarking}, 
which reveal that current models struggle with physical consistency across granular dimensions—ranging in complex scientific reasoning. 
To bridge the gap between visual realism and physical plausibility, researchers have proposed strategies such as neural Newtonian dynamics \cite{yuan2025newtongen}, progressive physical alignment \cite{wang2025prophy}, and verifiable rewards \cite{le2025gravity}, as well as VChain \cite{Vchain}, which enhances physical generation capabilities through instance-specific fine-tuning.
However, these methods typically rely on extensive physical datasets or require instance-specific test-time training.

\subsection{Unified Multimodal for Understanding and Generation}
The landscape of multimodal has been fundamentally reshaped by Large Multimodal Models such as GPT-4\cite{achiam2023gpt} and Gemini\cite{team2023gemini}. 
These foundation models have demonstrated unprecedented proficiency in both semantic reasoning and content generation, suggesting a convergence of perception and synthesis tasks. 
Inspired by this success, recent research has explored architectures that intrinsically unify understanding and generation within a single model. 
Specifically, Show-o\cite{showo} uses discrete tokens while BLIP3-o\cite{chen2025blip3} leverages continuous features to unify vision and language. Bagel\cite{bagel} further validates the emergent capabilities of such pretraining. Inspired by this success in static images, recent research is extending these unification paradigms to the temporal domain.
UniVid~\cite{luo2025univid} and UniVideo~\cite{wei2025univideo} propose unified frameworks that jointly understand and generate videos via shared representations for dynamic scenes.

However, constructing such understanding video generation models from scratch necessitates large-scale pretraining, incurring prohibitive computational costs and data demands.
To address this, we propose a training-free paradigm named \textit{PhyRPR}\. 
By orchestrating the reasoning capabilities of off-the-shelf multimodal models, a precise toolkit for coarse video generation, and diffusion models for visual refinement, our framework decouples understanding from generation in physical motion processes. 
This enables precise physics-constrained synthesis without expensive parameter optimization.

\section{Method}
\begin{figure*}[t]
    \centering
    \includegraphics[width=\textwidth]{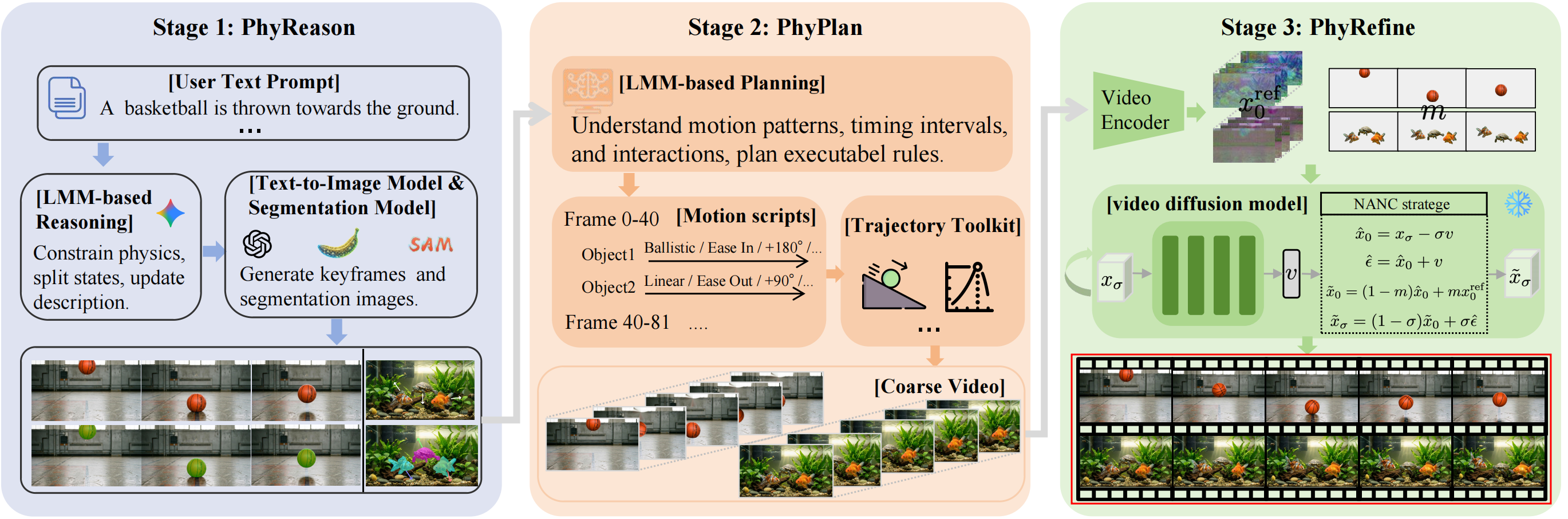}
    \caption{Overview of our training-free three-stage pipeline \textit{PhyRPR}: \textit{PhyReason}, \textit{PhyPlan}, and \textit{PhyRefine}.
    Stage~1 (see \ref{sec:phyreason}) outputs physically consistent keyframes and object states.
    Stage~2 (see \ref{sec:phyplan}) uses an LMM to select motion primitives and parameters, and deterministically renders a coarse motion video.
    Stage~3 (see \ref{sec:phyrefine}) applies motion-aware noise-consistent injection(NANC) to enforce the planned kinematics while preserving visual coherence.}
    
    \label{fig:pipeline}
    \end{figure*}

\subsection{\textit{PhyReason}: Visually Grounded Physical Reasoning}
\label{sec:phyreason}

To bridge the gap between textual physical understanding and video generation, the \textit{PhyReason} stage aims to obtain accurate visual representations of key physical states by leveraging
(i) a large multimodal model (LMM) for reasoning and (ii) a high-quality single-frame image generator for synthesis.

Given a user prompt $\mathcal{P}$, we first use the LMM to construct a physically constrained state-transition prompt sequence.
To encourage physically valid dynamics rather than merely narrative coherence, we employ prompt engineering to explicitly focus on key kinematic moments, including pivotal transitions and representative states over longer intervals, rather than uniformly sampling timestamps.
Formally, let $p_i$ denote the textual description of the physical state at the $i$-th kinematic milestone, and the entire sequence is jointly generated as:
\begin{equation}
    \{p_i\}_{i=1}^{L} = \mathrm{LMM}(\mathcal{P}).
    \end{equation}

While the textual state sequence provides a clear high-level scaffold, we further incorporate visual feedback to better align the state descriptions with concrete visual realizations.
Concretely, we first synthesize the initial keyframe $I_1$ from $p_1$.
For each subsequent milestone $i=2,\ldots,L$, the LMM takes the target state prompt $p_i$ as input and directly observes the previous keyframe $I_{i-1}$.
Based on this visual feedback, it produces a refined editing instruction $e_i$ that anchors the next step to salient visual attributes in $I_{i-1}$, and an instruction-guided image editing model generates the next keyframe:
\begin{equation}
I_i =
\begin{cases}
\mathcal{G}_{\mathrm{T2I}}(p_1) & \text{if } i=1, \\
\mathcal{G}_{\mathrm{edit}}(I_{i-1}, e_i) & \text{if } i>1.
\end{cases}
\end{equation}
where $\mathcal{G}_{\mathrm{T2I}}$ is the text-to-image generator and $\mathcal{G}_{\mathrm{edit}}$ is the instruction-guided editing model.

Finally, we apply an image segmentation model to parse the generated keyframes in an object-centric manner, extracting a binary mask $M_{i,k}$ for each dynamic entity $o_k$ at milestone $i$:
\begin{equation}
M_{i,k} = \mathcal{S}_{\mathrm{seg}}(I_i, o_k),
\end{equation}
where $\mathcal{S}_{\mathrm{seg}}$ denotes an open-vocabulary segmentation model that segments entity $o_k$ within frame $I_i$.

Overall, through physically grounded reasoning and closed-loop visual refinement, the \textit{PhyReason} stage produces a discrete set of physically self-consistent keyframes and object states, which serve as structured guidance for downstream planning and refinement.

\subsection{\textit{PhyPlan}: Physics-Aware Motion Planning}
\label{sec:phyplan}
Given the semantically consistent keyframes produced by \textit{PhyReason}, the \textit{PhyPlan} stage converts discrete object states into continuous spatiotemporal trajectories and synthesizes a motion-oriented coarse video that provides explicit kinematic guidance for diffusion refinement.

We treat a LMM as a motion director.
Given a keyframe sequence $\{I_i\}$, the corresponding object masks $\{M_{i,k}\}$, and optionally the original prompt $\mathcal{P}$, the model outputs a structured motion script that specifies per-entity motion semantics and a sequence of key states:
\begin{equation}
\mathcal{A} = \left\{ \left( \tau_k,\; \{\mathbf{s}_{i,k}\}_{i=1}^{L} \right) \;\middle|\; o_k \in \mathcal{O} \right\}.
\end{equation}
Here, $\mathcal{O}$ denotes the set of dynamic entities extracted in \textit{PhyReason}, $\tau_k$ is the motion primitive type of entity $o_k$ (e.g., \textit{Ballistic}, \textit{Drifting}, \textit{Linear}), 
and $\{\mathbf{s}_{i,k}\}_{i=1}^{L}$ represents a sequence of normalized state vectors. Each $\mathbf{s}_{i,k} = [x, y, s, r, \alpha]^{\top}$ encodes position, scale, rotation, and opacity at milestone $i$.

We implement a lightweight toolkit for trajectory synthesis.
For each neighboring pair of key states (between milestones $i$ and $i{+}1$), we instantiate the corresponding motion primitive $\mathcal{F}_{\tau_k}$.
We then fit its physical parameters $\theta^{*}$ to satisfy the boundary conditions and improve temporal coherence.
Let $\mathbf{c}_{\mathrm{start}}$ and $\mathbf{c}_{\mathrm{end}}$ denote the endpoint positions (e.g., object centers) at the two key states. The continuous trajectory is:
\begin{equation}
    \mathbf{c}(t) = \mathcal{F}_{\tau_k}\big(\mathbf{c}_{\mathrm{start}}, \mathbf{c}_{\mathrm{end}}, t;\, \theta^{*}\big),
\end{equation}
where $\mathbf{c}(t)$ denotes the continuous trajectory over time.
Our toolkit includes several common primitives sufficient for use.

Finally, we render the planned trajectories into a coarse visual signal via (i) layout synthesis and (ii) content composition.
Let $t \in \{1,\ldots,T\}$ index frames in the coarse video, and let $\tilde{\mathbf{s}}_{t,k}$ denote the per-frame state of entity $o_k$.
For each entity, we warp its initial mask $M_{1,k}$ and appearance crop $K_{1,k}$ according to $\tilde{\mathbf{s}}_{t,k}$, and obtain an occupancy map $\mathcal{M}_t$.
We then composite the objects onto the inpainted background $B^{(t)}$:
\begin{equation}
    \mathcal{O}_{t,k} = \mathrm{Trans}(K_{1,k}, \tilde{\mathbf{s}}_{t,k}) \odot \alpha_{t,k}\,\mathrm{Trans}(M_{1,k}, \tilde{\mathbf{s}}_{t,k}),
    \end{equation}
    \begin{equation}
    V_{\mathrm{coarse}}^{(t)} = B^{(t)} \odot (1 - \mathcal{M}_t) + \sum\nolimits_{k} \mathcal{O}_{t,k},
    \end{equation}
    
where $\mathrm{Trans}(\cdot)$ applies the geometric transformation specified by $\tilde{\mathbf{s}}_{t,k}$, $\alpha_{t,k}$ is its opacity component, and $\odot$ denotes element-wise multiplication.
The resulting coarse video $V_{\mathrm{coarse}}$ may exhibit stretched textures or imperfect details, but it preserves the intended topology and continuous trajectories, providing strong spatiotemporal guidance for latent-space refinement in the next stage.

In \textit{PhyPlan}, the planner selects motion primitives and parameters, and the toolkit deterministically renders a coarse video with physically plausible dynamics.

\subsection{\textit{PhyRefine}: Motion-Aware Visual Refinement}
\label{sec:phyrefine}

While the coarse video from \textit{PhyPlan} provides a physically grounded motion scaffold, it lacks fine-grained visual detail.
In \textit{PhyRefine}, we propose a motion-aware noise-consistent injection strategy that injects the scaffold as a latent-space constraint during sampling, leveraging the pretrained video model's rendering priors to improve motion adherence while refining appearance.

We derive the latent space motion mask $m$ by downsampling the occupancy masks $\{\mathcal{M}_t\}$ to the latent resolution and broadcasting it to match $x_\sigma$.
Let $x_0^{\mathrm{ref}}$ be the reference clean latent obtained by encoding the coarse video scaffold $V_{\mathrm{coarse}}$.
At selected sampling steps, given the current noisy latent $x_\sigma$ and the model-predicted velocity $v_\theta(x_\sigma,\sigma)$, 
we aim to enforce the planned state inside the mask while avoiding distribution shifts that may harm global appearance consistency.
We adopt an interpolation path for flow matching\cite{flow}:
\begin{equation}
x_\sigma = (1-\sigma)x_0 + \sigma \epsilon,
\label{eq:fm_param_local2}
\end{equation}
where $\sigma \in [0,1]$ is the noise level, $x_0$ is the underlying clean latent, and $\epsilon$ is the noise realization.
Under this path, the velocity field is defined by the derivative w.r.t.\ $\sigma$:
\begin{equation}
v^\ast \;=\; \frac{\partial x_\sigma}{\partial \sigma} \;=\; \epsilon - x_0.
\label{eq:fm_velocity_def}
\end{equation}
In practice, the video diffusion model predicts the velocity $v_\theta(x_\sigma,\sigma)$.
Thus, $\hat{x}_0$ and $\hat{\epsilon}$ below denote the \emph{implied} clean/noise components induced by $(x_\sigma, v_\theta)$ at the same $\sigma$:
\begin{equation}
\hat{x}_0 = x_\sigma - \sigma v_\theta, \qquad
\hat{\epsilon} = \hat{x}_0 + v_\theta = x_\sigma + (1-\sigma)v_\theta.
\label{eq:recover_x0_eps_local2}
\end{equation}
$\hat{x}_0$ reflects the model's current estimate of the clean video content, while $\hat{\epsilon}$ preserves the noise that governs textures, colors, and style along the current sampling trajectory.

We then inject the planned scaffold by replacing the clean component only within the motion region:
\begin{equation}
\tilde{x}_0 = (1-m)\hat{x}_0 + m x_0^{\mathrm{ref}}.
\label{eq:x0_inject_local2}
\end{equation}
This update aligns object content inside the mask with the planned scaffold, while leaving remaining regions unchanged.

Finally, we reconstruct a new latent at the same noise level $\sigma$ while explicitly preserving the recovered noise $\hat{\epsilon}$:
\begin{equation}
\tilde{x}_\sigma = (1-\sigma)\tilde{x}_0 + \sigma \hat{\epsilon}.
\label{eq:reconstruct_xt_local2}
\end{equation}
This step applies the physically grounded correction through $\tilde{x}_0$ but maintains the same noise manifold via $\hat{\epsilon}$.
The sampler proceeds with its update rule using $\tilde{x}_\sigma$ as the current state.
In practice, we apply this injection in early steps and continue sampling to obtain the final video.

In \textit{PhyRefine} stage, motion-aware noise-consistent injection strategy provides a simple training-free mechanism to couple \textit{PhyPlan}'s explicit kinematics with the video model's rendering capability, producing videos that are both trajectory-consistent and visually coherent.

\section{Experiments}

\begin{figure}[t]
    \centering
    \includegraphics[width=\columnwidth]{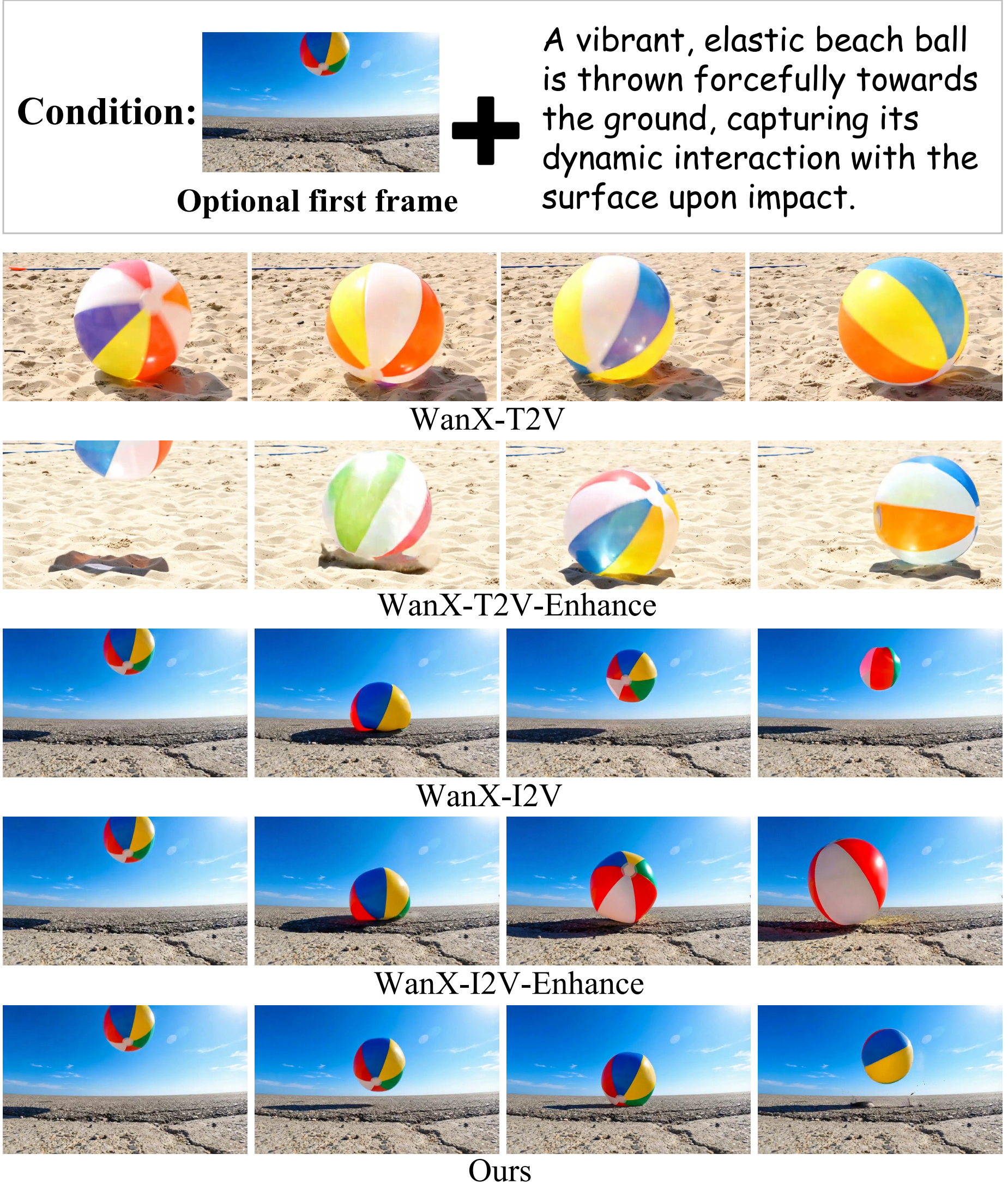}
    \caption{Qualitative comparison with baselines. We use the first frame from \textit{PhyReason} as the reference for I2V baselines. Our method better captures the physical process, including deformation during inflation and a plausible rebound.}
    \label{fig:baseline}
    \end{figure}

\begin{figure*}[t]
\centering
\includegraphics[width=\textwidth]{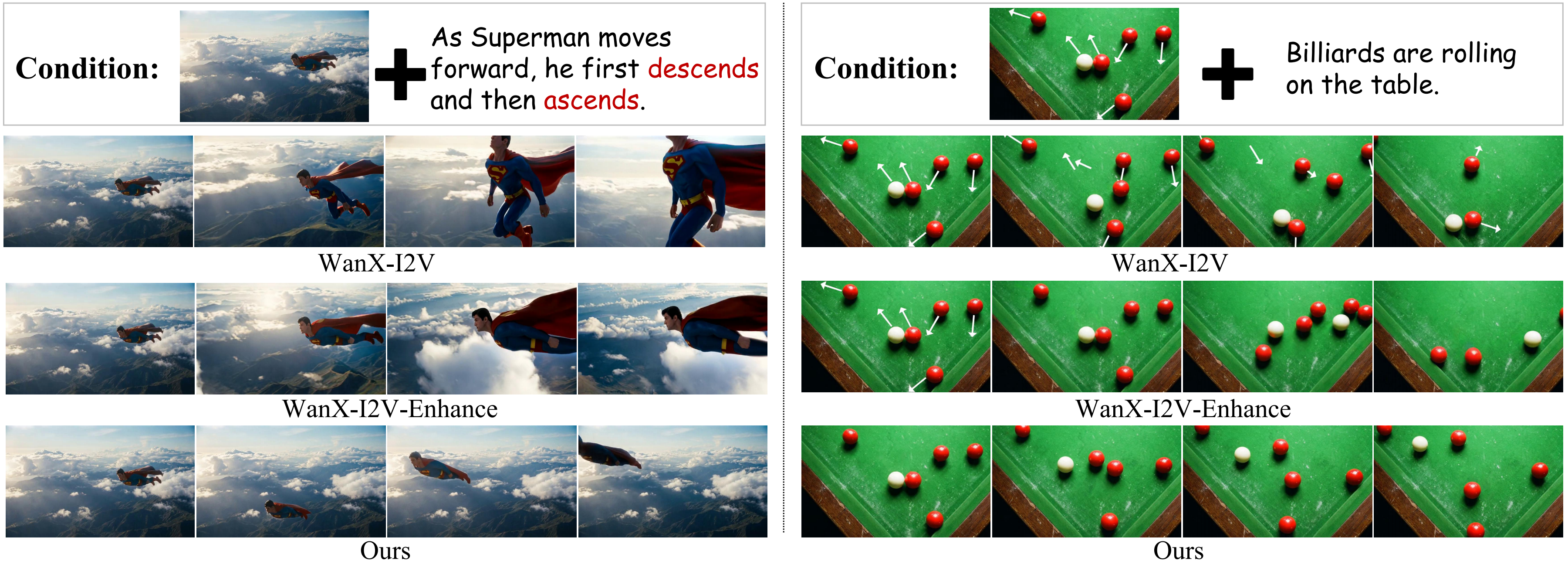}
\caption{Qualitative comparison with baselines. We specify physical constraints via text prompts or arrow guidance. Compared to baseline methods, our approach generates videos that more faithfully satisfy the specified physical constraints.}
\label{fig:baseline2}
\end{figure*}

\begin{table*}[ht]
  \centering
  \resizebox{\textwidth}{!}{%
    \begin{tabular}{lcccccccccccc} 
      \toprule
      \multirow{2}{*}{Method} &
      \multicolumn{3}{c}{VBench} &
      \multicolumn{5}{c}{LMM-as-judge (1-5)} &
      \multicolumn{4}{c}{User study (1-10)} \\
      \cmidrule(lr){2-4}\cmidrule(lr){5-9}\cmidrule(lr){10-13}
      & \makecell{Quality} & \makecell{Temporal} & \makecell{Overall}
      & \makecell{Physical\\plausibility} & \makecell{Trajectory\\compliance} & \makecell{Temporal\\consistency} & \makecell{Semantic\\alignment} & \makecell{Overall}
      & \makecell{Text\\alignment} & \makecell{Physics\\plausibility} & \makecell{Visual\\quality} & \makecell{Overall} \\
      \midrule
      WanX-T2V                 & 59.79 & 95.78 & 83.78 & 2.95 & 2.50 & 3.10 & 2.15 & 2.68 & 6.00 & 5.51 & 6.92 & 6.14 \\
      WanX-T2V-Enhance         & 54.23 & 91.94 & 79.37 & 3.45 & 3.05 & 3.73 & 3.23 & 3.36 & 6.45 & 5.55 & 6.82 & 6.27 \\
      \midrule
      WanX-I2V                 & 62.77 & 97.24 & 85.75 & 3.60 & 3.05 & 4.33 & 3.83 & 3.70 & 5.57 & 5.79 & 7.16 & 6.18 \\
      WanX-I2V-Enhance         & 60.88 & 95.26 & 83.80 & 3.85 & 3.63 & 4.45 & 4.13 & 4.01 & 6.18 & 5.99 & 7.49 & 6.56 \\
      LTX-Multi (-P)     & 59.59 & 97.54 & 84.89 & 3.65 & 3.88 & 4.10 & 4.00 & 3.91 & 6.32 & 5.86 & 6.02 & 6.07 \\
      SDEdit-style (-R)     & 61.66 & 97.71 & 85.70 & 3.55 & 3.43 & 4.38 & 3.95 & 3.83 & 6.13 & 6.01 & 6.93 & 6.35 \\
      \midrule
      \textit{PhyRPR} (ours)                     & 63.30 & 97.89 & 86.36 & 4.33 & 4.78 & 4.78 & 4.75 & 4.66 & 7.83 & 6.84 & 7.56 & 7.41 \\
      \bottomrule
    \end{tabular}%
  } 
  \caption{We evaluate with VBench, LMM-as-judge, and User study. \emph{-P,-R} denotes the ablation setting without \textsc{PhyPlan} and \textsc{PhyRefine}. Our method consistently outperforms baselines under physics constraints.}
  \label{tab:main_results}
\end{table*}

\subsection{Experimental Setup}

\textbf{Implementation Details.}
In \textit{PhyReason} and \textit{PhyPlan}, we use \texttt{gemini-3-pro-preview} as the LMM and SAM3~\cite{sam} for segmentation.
We synthesize milestone keyframes using \texttt{Nano Banana Pro}.
In \textit{PhyRefine}, we adopt \texttt{Wan2.2-I2V-A14B} as the diffusion model and perform training-free latent fusion during denoising. We evaluate under two common use cases of video diffusion models: text-conditioned generation and image-conditioned generation.
Under these settings, we design 40 diverse test scenarios spanning both text-only prompts and image+prompt pairs, and use the same set consistently for human evaluation, LMM-as-judge, and quantitative comparisons.
The I2V tasks includes both descriptive motion control and arrow-guided control.

\textbf{Baseline Settings.}
We use \texttt{Wan2.2-T2V-A14B} and \texttt{Wan2.2-I2V-A14B} as the base models for our baselines.
For fair comparison, each baseline is evaluated only under the protocol it supports.
Specifically, we report \texttt{Wan2.2-T2V-A14B} and its prompt-enhanced variant (\texttt{Wan2.2-T2V-Enhance}) on the T2V cases, and evaluate \texttt{Wan2.2-I2V-A14B} and its prompt-enhanced variant (\texttt{Wan2.2-I2V-Enhance}) on the all cases.
To control for the impact of prompt engineering, all ``-Enhance'' variants are produced by the same LMM-based rewriting procedure, which makes physical and kinematic constraints more explicit.
For I2V, the LMM takes both the reference first frame and the original prompt as input to ensure the rewritten motion description remains consistent with the visual context.

\textbf{Evaluation Metrics.}
We adopt three complementary evaluations to jointly measure overall video quality and physical Consistency.
First, we use VBench to quantify generic video quality aspects such as visual fidelity and temporal consistency.
However, these metrics are often insufficient to determine whether a model strictly follows physics-aware constraints.
We employ Gemini as an LMM-as-judge to assess constraint satisfaction along four axes, each scored on a 1--5 scale.
\emph{Physical plausibility} evaluates whether behaviors and interactions obey the physical or logical rules implied by the prompt, avoiding implausible artifacts.
\emph{Trajectory compliance} measures how accurately the video follows the specified motion path, directions, and event order.
\emph{Temporal consistency} checks object permanence over time, penalizing flicker, unprompted morphing, or disappearance.
\emph{Semantic alignment} assesses whether the video matches both the explicit prompt content and its implied requirements.
Finally, we conduct a user study with 12 participants, who rate each video on a 1--10 scale for text alignment, physical plausibility, and visual quality.

\subsection{Comparison with Baselines}
\textbf{Quantitative Comparisons.}
Table~\ref{tab:main_results} reports quantitative results with baselines.
Overall, I2V baselines achieve higher VBench scores than T2V baselines, which is expected since I2V is anchored by a reference first frame and thus benefits from stronger appearance initialization.
Prompt enhancement improves constraint following: while ``-Enhance'' may slightly reduce VBench quality, it boosts the LMM-as-judge and user-study scores, indicating better satisfaction of physical and motion constraints. In contrast, our method achieves the best performance across all reported metrics.

\textbf{Qualitative Comparisons.}
We present qualitative comparisons against baseline methods in Fig.~\ref{fig:baseline} and Fig.~\ref{fig:baseline2}.
In Fig.~\ref{fig:baseline}, the task requires the model to respect basic physical dynamics, including impact deformation and rebound.
T2V baselines generated directly from the prompt fail to produce a clear and physically plausible bounce after the volleyball contacts the ground.
I2V baselines conditioned on the same first frame as ours still exhibit implausible artifacts, including excessive deformation of the volleyball in the second column.
In contrast, our method produces both reasonable impact deformation and a coherent rebound consistent with the prompt.
Fig.~\ref{fig:baseline2} compares methods in physically constrained motion control scenarios.
The left example evaluates prompt-specified trajectories: baselines drift from the intended path, with or without prompt enhancement, whereas our method follows the described trajectory more faithfully.
The right example evaluates arrow-guided control: baselines often confuse directions across arrows and sometimes render unstable arrow cues over time, leading to chaotic motion.
Our method moves each billiards along its assigned arrow trajectory with better temporal coherence and physical plausibility.

\subsection{Ablation Studies}
We report two ablations in Table~\ref{tab:main_results}: SDEdit-style (\emph{-R}) and LTX-Multi (\emph{-P,-R}).
Here \emph{-P} removes \textit{PhyPlan} and \emph{-R} removes \textit{PhyRefine}.
For controlled comparison, all variants share the same \textit{PhyReason} outputs, which provide the structured inputs required by downstream planning and refinement.

\textbf{Ablation on \textit{PhyPlan}.}
We use LTX-Multi\cite{ltx}, a strong video generation foundation model that can take multiple input images together with their target frame indices.

\textbf{Ablation on \textit{PhyRefine}.}
To isolate the effect of \textit{PhyRefine}, which uses motion-aware noise-consistent injection to enforce the planned scaffold during denoising, we replace it with a standard SDEdit-style alternative~\cite{sdedit}.
Specifically, we add noise to the entire coarse video $V_{\mathrm{coarse}}$ to a fixed noise level and then denoise it with the same video diffusion model.
This setting performs global refinement but does not selectively constrain the planned motion regions, serving as a strong training-free method.

\section{Conclusion}
In this work, we introduce a training-free three-stage pipeline \textit{PhyRPR} for physics-constrained video generation. By decoupling physical understanding from visual synthesis, we infer key physical states, plan a motion scaffold, and inject it into diffusion sampling via noise-consistent latent fusion. Experiments show gains in physical plausibility and motion controllability while preserving high visual quality.
\bibliographystyle{IEEEbib}
\bibliography{PhyRPRreferences}

\vspace{12pt}

\end{document}